\def\year{2017}\relax
\documentclass[letterpaper]{article} 
\usepackage{aaai17}  
\usepackage{times}  
\usepackage{helvet}  
\usepackage{courier}  
\usepackage{url}  
\usepackage{graphicx}  
\usepackage{multirow}
\usepackage{color}
\usepackage{blindtext}
\usepackage[inline]{enumitem}
\usepackage{xcolor}
\usepackage{tabularx}
\usepackage{afterpage}
\usepackage{capt-of}
\usepackage{mwe}  
\usepackage{subcaption}
\usepackage{tabularx}
\usepackage{amsmath} 

\begin{document}
%
\title{Fashion Conversation Data on Instagram}
\author{Yu-I Ha$^*$~~~~~~~~~Sejeong Kwon$^*$~~~~~~~~~Meeyoung Cha$^*$~~~~~~~~~Jungseock Joo$^{\dag}$\\
~\\
$^*$Graduate School of Culture Technology, KAIST, South Korea\\
$^{\dag}$Department of Communication Studies, UCLA, USA\\
}
\maketitle
\begin{abstract} 
\begin{quote}
The fashion industry is establishing its presence on a number of visual-centric social media like Instagram. This creates an interesting clash as fashion brands that have traditionally practiced highly creative and editorialized image marketing now have to engage with people on the platform that epitomizes impromptu, realtime conversation. What kinds of fashion images do brands and individuals share and what are the types of visual features that attract likes and comments? In this research, we take both quantitative and qualitative approaches to answer these questions. We analyze visual features of fashion posts first via manual tagging and then via training on convolutional neural networks. The classified images were examined across four types of fashion brands: mega couture, small couture, designers, and high street. We find that while product-only images make up the majority of fashion conversation in terms of volume, body snaps and face images that portray fashion items more naturally tend to receive a larger number of likes and comments by the audience. Our findings bring insights into building an automated tool for classifying or generating influential fashion information. We make our novel dataset of {24,752} labeled images on fashion conversations, containing visual and textual cues, available for the research community.
\end{quote}
\end{abstract}

\noindent Social media has become an important platform for the fashion industry for testing new marketing strategies and monitoring trends~\cite{kim2012social}. Already thousands of luxury and high street fashion brands around the world are present online and  communicate with their followers and potential customers~\cite{hu2014we}. While fashion brands have unilaterally set their polished brand images through traditional media such as television channels and magazines, two unique properties of social media serve as a very powerful tool for promoting and sharing fashion information to both industry and people. 

Firstly, the interactive nature of social media allows anyone to generate content and participate in establishing brand images. Not only large fashion houses launch advertising campaigns and share their latest runway looks through social media platforms, individuals and local stores also contribute to the online fashion conversation by sharing purchase experiences or new trends. Individuals discuss and rate fashion products openly and favorable reviews spread via word-of-mouth. In contrast, dissatisfied customers leave harsh criticism and complaints on fashion products online.


Secondly, many social media platforms are visual-centric and heavily utilize images and videos. Visuals, a powerful tool in advertising and communication~\cite{messaris1996visual,joo2014visual}, are critical in fashion marketing because appearance  is the key information of any fashion look. In addition, compared to traditional platforms, social media offer much higher bandwidth in that brands can now deliver information about a single fashion product through hundreds of varying images. 

Among various platforms, Instagram has reshaped the fashion industry landscape. 
Plenty of fashion brands are hosting new marketing campaigns based on it's hashtag functions and consumer-generated contents. To meet the spontaneous nature of the platform, some brands also adopt non-traditional photographic styles such as ``behind the scenes'' that are secondary, less-editorialized images to create unique brand stories.  

However, little is known about the fashion conversation itself. Lack of any labeled data describing fashion style is a barrier to investigating such trends. Building a comprehensive fashion dataset will enable research on customized recommendations that associate personal tastes with fashion picks as well as identify emerging trends from different parts of the world. Such research will enable fashion industry to better understand how products and brands are perceived by people. As a result building fashion datasets helps create new products in a sustainable way. According to the World Economic Forum, fashion is the second largest polluter of environment after oil and creating innovative ways to reduce fashion waste is a critical challenge. 


We envision to take on the challenge of analyzing how fashion tastes are shared and disseminated on social media. As a first step, we built a sizable yet detailed labeled dataset describing conversations on notable fashion brands on Instagram. By employing deep learning techniques, we identified meaningful topics in the context of fashion and automatically labeled the gathered images. This paper presents the steps involved in the labeling task and shares the data for further discussion. Our contributions are as follows:
 
\begin{enumerate}


\item We release a novel dataset describing 24,752 fashion images of 48 brands on Instagram with meaningful visual tags. We identified  five major visual categories of fashion images via comprehensive content labeling, which are selfie, body snap, marketing shot, product-only, and non-fashion. We also trained a convolutional neural network (CNN) to classify the major visual categories and other important visual features from images including `face' or `brand logo.'

\item Our analysis on visual content of fashion images and audience engagement reveals an interesting discrepancy between post volume and reactions; while product-only images are the most common in terms of volume, body snaps and photos containing faces that reveal fashion items more naturally receive disproportionately large number of likes and comments from the audience (e.g., 31\% of the fashion posts receiving 53\% of total likes). 

\item We identify challenges in fashion image classification; we encounter a number of non-canonical images that do not fall into the five major categories such as advertisements, clickbaits, zoom-in shots of textile and products, and multi-functional images. These images nonetheless explicitly contained  fashion hashtags.  

\item Regression and ANOVA tests indicate what kinds of image features and emotions draw more attention from audience. We find that body snap and face features are a better communicator than product-only or logo features and that certain facial expressions like happiness and neutral emotion show a significant relationship with the likes count.


\end{enumerate}  

Our findings bring theoretical and practical implications for studying fashion conversations on social media. The tagged images can be used for defining what images are considered in the domain of fashion in user generated content. For example, the practice of sharing images containing faces with fashion hashtags needs to be understood better. Some of these images dedicated more visual space to the face itself than the associated fashion items (i.e., selfie shots where faces take up more than half of photo length). We also observed a number of image spams that exploit irrelevant fashion tags, and hence our data can be potentially used for identifying fashion clickbaits. Finally our data indicate that product-only images dominate the conversation of most fashion brands, yet images of this type are less effective in gaining likes and comments. Such information will be useful for fashion brands and marketers to effectively promote their products and communicate their brand images to the customers.

\section{Related Work} 

Online social media have transformed the way in which many fashion brands promote and sell their products, establish brand identities, and increase customer loyalty. For successful brand promotions, the fashion industry has adopted various online marketing strategies, which operate in social media platforms such as finding potential customers and tracking a list of users by specific events~\cite{du2013social}. In addition, many fashion brands have brought their online shopping functions into existing social media sites. For example, Levi's, Lands' End, and Express allow consumers to buy their apparel products from their Facebook pages~\cite{kang2012social}. This greatly helps fashion brands understand their target customers and their preferences, thereby enabling various targeted marketing strategies as well. 

Many prior studies have investigated the effects of marketing in social media on various consumer behaviors and brand perceptions. Online social media and product review forums are extensively used to express views and share experiences about products~\cite{jhamtani2015identifying}. Consumers on social media actively communicate with each other and with companies~\cite{dijkmans2015stage}. They are also knowledgeable about brands and their products~\cite{agnihotri2016social}. In the domain of fashion, marketing activities of luxury fashion brands on social media have been shown to be positively correlated with eventual purchase intention of customers, mediated by brand perception~\cite{kim2012social}. Consumers actively collect and utilize information from both internal sources (e.g., brand familiarity, prior shopping experience) and external sources (e.g., web sites) ~\cite{chiang2006clicking}. Purchasing decisions can be affected by various factors such as opinions from their friends and relatives ~\cite{qian2015study}, the visual images of products, and satisfactory information~\cite{hye2002apparel}. Therefore, social media which host the sheer amount of information about brands and products are an ideal place for customers to gather necessary information and make decisions.

Instagram specializes in instant communication of trends and visual information in general, in a way similar to Twitter, Pinterest, and Flickr~\cite{park2016style}. It has rapidly grown in the recent years and the fashion brands' engagement has far outstripped the audience growth rate. Although the most socially active brands have an Instagram account, fashion related brands which were classified as luxury retail, clothing, beauty and consumer merchandise have strong presence on this platform. 
In addition, as not only attracting fashion bloggers, but also generating new fashion-related micro-celebrities, fashion-related postings and conversations came pouring out every day on Instagram. 

The fashion industry has focused on increasing brand equity via creating new collaborations and interacting with their followers on social media. The fashion show, one of the most representative offline events in the fashion industry merges into Instagram by encouraging public users to engage their hashtag campaign during fashion weeks. As fashion-related publications such as magazine, look book, and even video of runway show have moved online social media, consumers utilize social media platform as they read magazines. This phenomenon has changed the fashion landscape and the users' information acquisition method. The existing studies, however, have not considered the effects of visual marketing on social media with respect to the contents of images. This is important to understand because different visual portrayals of products can invoke distinct responses, which may lead to different effects on purchase decision.


\begin{figure*}[t!]
	\centering        \includegraphics[width=1\textwidth]{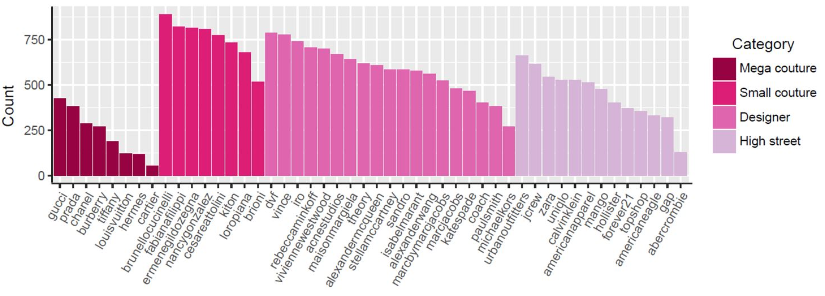}
    \vspace*{-10mm}
	\caption{Post frequencies of the final 48 fashion brand data } 
	\label{fig.hist.brand}
\end{figure*}

\section{Data Labeling}



We collected Instagram posts and collected engagement logs per post such as the number of likes and comments. In addition to these features, we added  visually meaningful tags such as facial emotion, brand logo, and the number of faces based on deep learning models.

\subsection{Data Collection} 

Fashion posts were accessed through the Instagram's API over a week period in January, 2015. We searched for posts mentioning particular hashtags such as keywords describing fashion in general (e.g., fashion, ootd) as well as specific brand names. General terms such as `fashion' or outfit-of-the-day (ootd) were used in many other contexts beyond fashion. Hence, for this work, we limit to a set of posts containing at least one hashtag of a brand (e.g., \#gucci). The brand list contained 48 internationally renowned names from luxury fashion houses like Hermes, Prada, and Louis Vuitton to high street brands like Zara and Forever 21. Our criteria for choosing a brand name was (1) whether they have a strong presence on Instagram (i.e,. have an official account with at least a 5,000 followers) and (2) there are enough individuals mentioning the brand (e.g., more than 7,000 posts). The full list of brand names and their grouping appears on Figure~\ref{fig.hist.brand}, where brands belong to four groups --- mega couture, small couture, designer, and high street --- based on brand identity, popularity, and price ranges. 


The initial set of fashion posts (gathered over 50,000 instances) were newly uploaded content at the time of crawling. To examine the likes and comments of these posts, we re-gathered information about these fashion posts from the API after a week time, which is considered enough time period to gather sizable reaction from audience in streaming-based social media~\cite{szabo2010predicting}. 
Some posts have been deleted or have been made no longer available, in which case were excluded. The final dataset for which we have complete information about the uploader, the image itself, and the text description were 24,752 fashion photo posts uploaded by 13,350 distinct individuals. For each post, we obtained the following information: 




\begin{itemize}
\item User Id: Numeric ID of the posting user 
\item User name: Screen name of the posting user
\item User profile picture URL: A web link to the user's profile picture of the post
\item Followings: The number other users a given uploader is following (i.e., distinct sources of the uploader)
\item Followers: The number of other users subscribing to the uploader's account (i.e., fans of the uploader)
\item Media count: The number of total posts (both fashion and non-fashion) contributed by the uploader
\item Brand name: A brand name used in fashion post search process, used as a hashtag in user's post
\item Brand category: Grouping of a brand 
\item Hashtags: The list of hashtags in a post 
\item Caption: Text description of a post provided by the user, excluding the hashtag information
\item Image URL: A web link to the image file of the post 
\item Likes: The total number of likes per post as well as the list of Instagram audience who liked each photo (including their user IDs, names, and profile URLs) 
\item Comments: The total number of comments as well as the list of Instagram audience who commented on the photo
\item Creation Time: When post was uploaded on Instagram 
\item Link: A web URL (if any) contained in each post
\end{itemize}

\begin{figure}[t!]
	\centering        \includegraphics[width=0.45\textwidth]{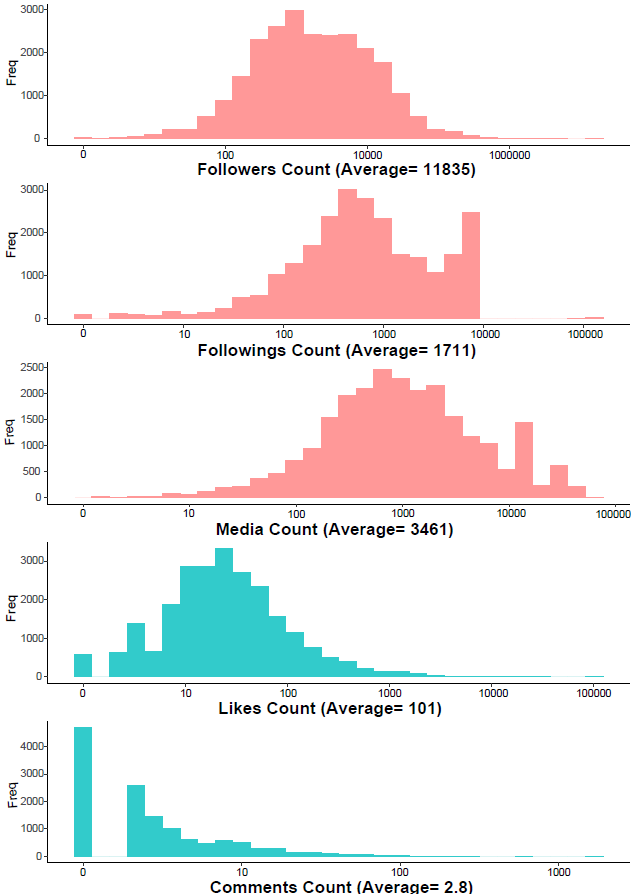}
	\caption{Popularity and activity distribution of uploaders and posted content}
	\label{Hist}
\end{figure}

Figure~\ref{Hist} shows the followers, followings, and media count histograms of the fashion post uploaders as well as the distribution of likes and comments on such content. The x-axis represents the value of the given feature, and the y-axis represents frequency. The high average values denoted in the x-axis indicate that individuals posting fashion information on Instagram likely have more followers, followings, and media counts than ordinary users. The likes count also shows a peak at 6, indicating that fashion posts are less likely to be completely ignored (i.e., receiving zero likes). Comments, however, is a more effort-taking reaction and the largest fraction of posts received zero comments. 


\subsection{Labeling Images with Deep Learning}
To characterize visual features and content in fashion images, we automatically classify images and generate tags by a convolutional neural network (CNN). Recent studies in computer vision have proposed approaches to automatically analyze fashion images. The traditional line of research focuses on classifying individual fashion items such as `jacket' or `t-shirt' in images \cite{bourdev2011describing,bossard2012apparel,vittayakorn2015runway,liu2016deepfashion}. Based on such techniques, other studies also attempt to recognize the overall styles \cite{kiapour2014hipster}, perceptual fashionability and fashion trends in image datasets \cite{simo2015neuroaesthetics}. A few more studies further analyze fashion content to investigate the marketing strategies of fashion brands \cite{manikonda2015trending,chen2016fashion}. While we also share some similar motivations with these studies, our paper and dataset explicitly examine how images are ``framed'' (e.g., selfie, body snap, etc) and its relations to user perception and brand marketing on Instagram. Our automated labeling is therefore aimed at gaining more insights on the influential roles of fashion posts to information seeking and decision making of users by quantifying visual features.


\subsubsection{Defining Visual Categories}


We first define visual content variables of fashion posts that draw attention from Instagram audience. While fashion posts commonly display fashion items, many other visual cues can affect the effectiveness of an image, such as presence of a fashion product with a face, body part(s), or a brand logo as well as the emotion of the main subject. To semantically quantify visual content of fashion images, we relied on the grounded theory approach using manual content tagging \cite{cohn1999automated}. 

\begin{figure}[t!]
	\centering
	\includegraphics[width=0.48\textwidth]{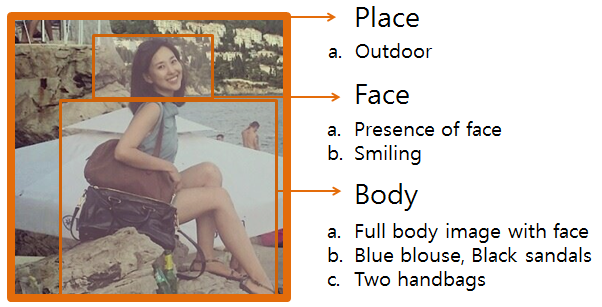} \\
	(a) Image tagging process \\
	~\\
	\includegraphics[width=0.48\textwidth]{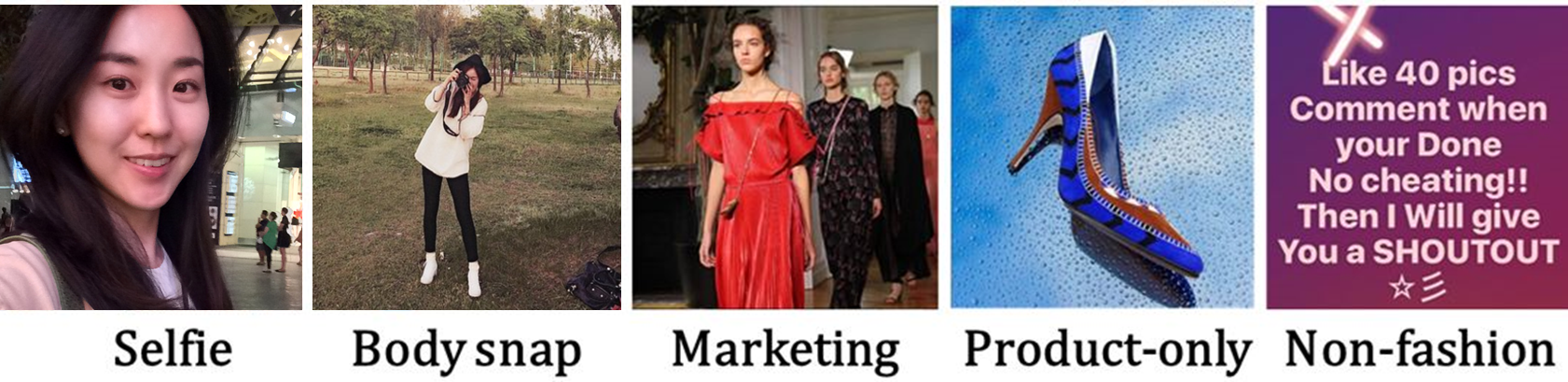}\\
	\vspace*{1mm}
	(b) Fashion image categories and examples
	\caption{Image categorization process, where  images were identified by themes, merged into similar concepts, and through iterative steps were combined as five categories. }
	\label{fig.tagging}
\end{figure}

As the first step, we manually tagged visual information of 1,000 sample data by identifying themes and repeatedly merging them into similar concepts until a handful of categories emerged. Manual tagging was conducted by the first author, who has majored in digital fashion. While tagging process on the sample images was conducted by one person, the process of building major themes and deciding criteria for clustering was done along with a domain expert, who had been in the fashion industry for ten years.

\begin{table*}[t!]
	\centering 
	\caption{The list of visual content variables in our analysis and descriptions}
	\label{tab.imagefeature}
	\begin{tabular}{cccl}
		\hline 
        Type 
        & Variable &Quantity & Description \\ \hline 
        Major category & Selfie &Prob & Images whose face occupies more than 50\% of height  \\ 
		 & Body snap &Prob &Images including full or partial body parts and not a selfie  \\ 
		& Marketing &Prob & Editorialized images like runways, awards ceremonies  \\  
		& Product-only &Prob & Images without face or body parts that contain fashion products  \\  
        & Non-fashion &Prob & Images not related to fashion such as food, landscape, and animals\\ \hline
		Subcategories &Face  &Prob & Images containing frontal or side faces   \\ 
		& Logo &Prob & Images with unmatched brand logo or text   \\ 
        & Brand logo  &Prob & Images with the matching logo of the brand in hashtags  \\ 
        & Smile &Prob  & Images containing one or more smiling face(s) \\ 
		& Outdoor &Prob  & Images whose background is outdoor \\ 
		& People &Count & The number of people detected in the image via body shapes \\ 
		& Items &Count & The number of fashion items estimated in the image \\  \hline                         
	\end{tabular}
\end{table*}

In the first round of tagging, images received detailed tags such as `smiling face' or `face with eyes closed,' which were merged as `selfies.' The images that mention fashion keywords explicitly yet are not directly related to brand or fashion were classified as `non-fashion.' In the next iterative tagging rounds, similar tags were merged into a single tag. After several rounds, we arrived at five visual categories and seven image features that compose the studied fashion images. 

The five categories are listed in Figure~\ref{fig.tagging}. The most popular category was photos of body snaps containing mostly half or full body images, which made up 23\% of the samples. We also included images with body parts without necessarily revealing torso (e.g., holding a handbag) as body snap. The second most popular category was marketing shots (22\%), which were runway or advertisement images produced by professionals. The biggest difference between body snaps and marketing shots was the image composition, for example, whether the image composed with the studio setting such as light, background, and presence of professional model, celebrities or not. The third category was product-only (20\%), which contained fashion items related to the mentioned brands without any human body. The marketing shots and product-only types were the most common kinds photo that official brand accounts shared. 

The next set of images were classified as selfies (18\%), which majorly contained faces and its surrounding areas. Fashion items could not be seen clearly in these images. We measured the face length to determine selfies via assuming selfie faces take up more than half of photo height. The final type was a mix of food, landscape, indiscernible products, and text-based memes, which we classified as non-fashion (17\%). The tagging process and example of image category is depicted in Figure~\ref{fig.tagging}.\footnote{All photographs shown in this paper are public content. We chose photos shared by popular individuals on Instagram, who have at least 100 followers. We obtained permissions from the authors for personal photographs (e.g., selfie) } Finally we arrived at 5 major visual categories and 7 subcategories, as summarized in Table~\ref{tab.imagefeature}. We allowed multi-label classification and each photo could belong to more than one major- and sub- categories. In total, 71.3\% of photos were classified as one of the five major categories and a great majority of them (48.7\%) had a single label. Variables are in the form of probability, except for People and Products counts.

\subsubsection{Classification with CNN}

To automatically classify the image features specific to the five meaningful fashion categories, we trained a convolutional neural network (CNN) using the annotations of 3,169 fashion images as a training set. Our model architecture is Residual Network of 50 layers (ResNet) \cite{he2016deep}, which has demonstrated the state-of-the-art performance in image classification. ResNets utilize ``skip'' connections to directly leverage features from the lower layers of CNNs and combine them with more structured features from the higher layers. 

The task is posed as a multi-label classification in which each output is treated separately. We combine the independent binary cross entropy losses and the mean squared errors to cope with 10 binary classes and 2 integer classes. To leverage generic visual features, we used a public model pretrained on Imagenet data,\footnote{http://torch.ch/blog/2016/02/04/resnets.html} replaced the last layer, and fine-tuned the model with our own data. We begun with a learning rate of 0.001 and decreased it to 0.0001 after 10 epochs with a weight decay rate of 0.0005. We applied standard data jittering (translation, scaling, color variation) to augment the training set to avoid overfitting. 


\begin{table}[t!]
	\centering
	\caption{The result of image classification. MSE represents the mean squared errors }
	\label{imageclassification}
	\begin{tabular}{clcc}
		\hline
		&\multicolumn{1}{c}{\multirow{1}{*}{\begin{tabular}[c]{@{}c@{}}Classified\\posts (\%)\end{tabular}}} 
		&\multicolumn{1}{c}{\multirow{1}{*}{\begin{tabular}[c]{@{}c@{}}Avg \\ Precision \end{tabular}}} &
		\multicolumn{1}{c}{\multirow{1}{*}{\begin{tabular}[c]{@{}c@{}}Positive\\ratio \end{tabular}}} \\ 
		\\ \hline                     
		Selfie     &336 (1.4\%) & 0.90   & 0.03   \\                                                       
		Body snap  &5582 (22.5\%) & 0.90   & 0.27   \\                                                   
		Marketing  &1298 (5.2\%)& 0.77   & 0.10   \\                                                    
		Product-Only    &8588 (34.6\%) & 0.97   & 0.44   \\                                                    
		Non-fashion &1873 (7.5\%) & 0.88   & 0.14   \\                                                     
		Face       &6936 (28.02\%) & 0.94   & 0.30   \\                                                   
		Logo       &8856 (35.8\%) & 0.93   & 0.52   \\                                                     
		Brand logo &317 (1.3\%) & 0.42   & 0.16   \\                                                     
		Smile      &148 (0.6\%) & 0.59   & 0.08   \\                                                       
		Outdoor   &1465 (5.9\%) & 0.71   & 0.11   \\  \hline                                                  
		&\multicolumn{1}{c}{\multirow{2}{*}{\begin{tabular}[c]{@{}c@{}}Avg~count\end{tabular}}} &\multicolumn{1}{c}{\multirow{2}{*}{\begin{tabular}[c]{@{}c@{}}MSE\end{tabular}}} &\multicolumn{1}{c}{\multirow{2}{*}{\begin{tabular}[c]{@{}c@{}} $R^2$ \end{tabular}}} \\ 
   
		\\ \hline                         
		\# of People & 0.7418518 & 0.8557   & 0.3678   \\  
		\# of Items & 3.261685 & 4.0457  & 0.4649  \\ \hline
	\end{tabular}
    \vspace*{-5mm}
\end{table}

\begin{table*}[!h]
	\centering
	\caption{Examples of successful visual content classification } 
\label{classification_failure}
	\label{classification_success}
	\begin{tabular}{rlrrrrrrrrrrrrr}
		\multicolumn{10}{c}{\includegraphics[width=0.9\textwidth]{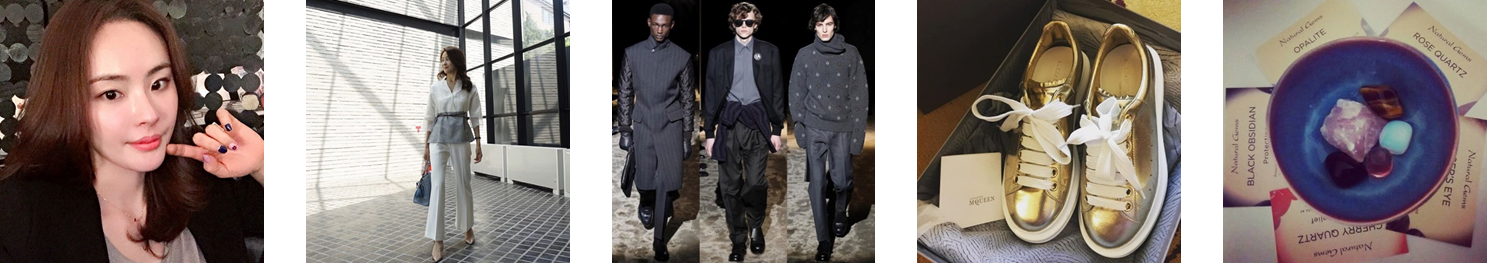}}\\
        
		\multicolumn{10}{c}{~~~~~~Image 1 (Selfie)~~~~~~~~~~~~~~Image 2 (Body snap)~~~~~~~Image 3 (Marketing)~~~~~~~Image 4 (Product)~~~~~~~Image 5 (Non-fashion)}\\  ~\\ \hline
        
        \end{tabular}
        \begin{tabular}{cccccccccccccc}
		& Selfie & Body snap & Marketing & Product-only &Non-fashion& Face & Logo & Brand logo& Smile&  Outdoor\\ \hline 
		Image 1& \textbf{0.868}&0.398 &0.016 &0.002 &0.013 &0.982 &0.068 &0.020 &0.414 &0.009 \\ 
		Image 2& 0.002 &\textbf{0.999} &0.003 &0.000 &0.000 &0.884 &0.024 &0.003 &0.188 &0.472 \\ 
		Image 3& 0.002 & 0.119 &\textbf{0.862} &0.016 & 0.004 & 0.936 & 0.028 & 0.024 &0.058 &0.105 \\ 
		Image 4& 0.000 & 0.000 &0.000 &\textbf{0.999} &0.000& 0.000& 0.997 &0.480 &0.000& 0.000 \\  
        Image 5& 0.002 & 0.005 &0.006 &0.025 &\textbf{0.993} &0.006 &0.997 &0.121 &0.006 &0.001 \\ 		
		\hline
	\end{tabular}
\end{table*}

\begin{table*}[!h]
	\centering
	\caption{Examples of unsuccessful visual content classification }
	\label{classification_failure}
	\begin{tabular}{rlrrrrrrrrrrrrr}
		
		\multicolumn{10}{c}{\includegraphics[width=0.9\textwidth]{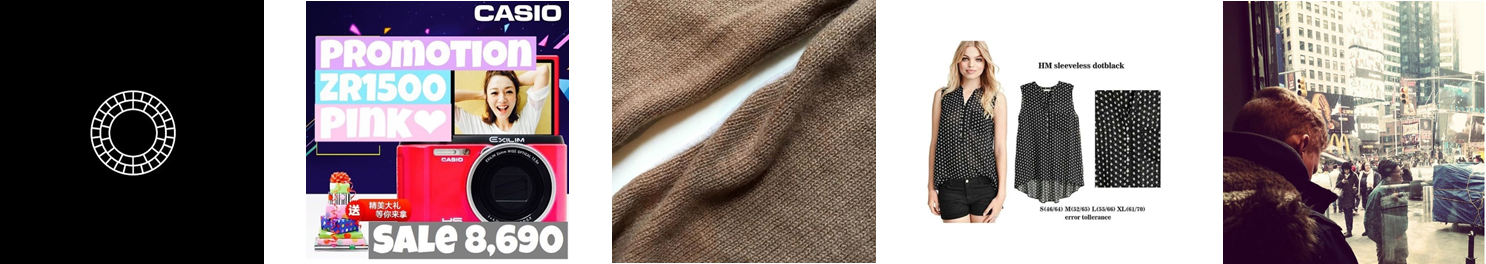}}\\
		\multicolumn{10}{c}{Image 1 (Others)~~~~~Image 2 (Advertisement)~~~~Image 3 (Textile)~~~~~Image 4 (Marketing)~~~~~~Image 5 (Multifunctional)}\\  ~\\  \hline
        \end{tabular}
        \begin{tabular}{ccccccccccccc}
		& Selfie & Body snap & Marketing & Product-only &Non-fashion& Face & Logo & Brand logo& Smile&  Outdoor\\ \hline 
		Image 1& 0.002 & 0.013 & 0.032 & 0.256 & 0.474 & 0.023 & 0.289 & 0.042 & 0.003 & 0.013\\ 
		Image 2& 0.019 & 0.074 & 0.204 & 0.016 & 0.444 & 0.878 & 0.995 & 0.162 & 0.268 & 0.018\\ 
		Image 3& 0.007 & 0.130 & 0.012 & 0.190 & 0.179 & 0.013 & 0.057 & 0.198 & 0.004 & 0.092\\  
		Image 4& 0.002 & 0.061 & 0.573 & 0.045 & 0.028 & 0.516 & 0.804 & 0.224 & 0.037 & 0.026\\  
		Image 5& 0.068 & 0.272 & 0.034 & 0.037 & 0.550  & 0.383 & 0.249 & 0.085 & 0.099 &	0.106\\ 
		\hline
	\end{tabular}
    \vspace*{-3mm}
\end{table*}

We measured the classification accuracy by average precision (AP) for 10 binary variables, and by the mean squared error for 2 integer classes (number of people or products), as shown in Table~\ref{imageclassification}. AP is the most common measure for binary classification in computer vision tasks. The table also shows the portions of image posts classified as positives by each visual variable. Based on the classification result, we examined the successful and failed cases in Table~\ref{classification_success} and Table~\ref{classification_failure}. 
The numbers of people and fashion items were not directly used in our following analysis but included in our dataset for future research.

The results indicate that our trained classifiers are highly effective in classifying visual variables and thus provide reliable annotations for visual information of images. The AP values for all variables are much higher than their positive example ratios, which are the expected AP values for random or majority classifiers. For integer classes counting the number of people or items in an image, our model directly outputs the predicted values with reasonable accuracy. An alternative approach would be to detect object instances separately; however, this requires much heavier computations and a separate pipeline from binary classification. Therefore, we use a 1-step holistic CNN for integrity and simplicity.

 
We present a few concrete examples in order to provide insights on the model performance. Table~\ref{classification_success} shows a few common examples in our dataset and their prediction scores that correctly describe the visual features. In contrast, Table~\ref{classification_failure} shows a few challenging cases in which the images are not strongly associated to any of the five major categories. 
In fact, images such as 1 (synthetic) and 2 (ads with overlaid text) may be deemed irrelevant in the context of fashion. Since these images include fashion brand hashtags, they may be considered clickbaits. Image 3, however, is a zoom-in shot of textile and may be considered as meaningful fashion information. Image 4 may be considered a marketing shot, yet its overlaid text makes it hard to distinguish from other clickbaits. Finally image 5 is a multi-functional type that has no clear type (even to human judges its relevance to fashion is not clear). These failure cases, which are also included in the data along with CNN classification results, pose challenges in determining what kinds of information should be considered relevant to fashion conversation.

\begin{table*}[!h]
\centering
\caption{Meta analysis results about users' engagement and fashion image category}
 
\label{features_engagement}
\begin{tabular}{crrrrr}
\multicolumn{1}{c}{\multirow{2}{*}{\begin{tabular}[c]{@{}c@{}} \textsf{Aggregate Data} \\ (N=24752, 71\% classified)  \end{tabular}}}\hspace{\parindent} &
\\
& Selfie & Body snap & Marketing & Product-Only & Non-fashion  \\ \hline
Total posts       & 336 (1.9\%)   			  & 5582 (31.6\%)			 & 1298	(7.3\%) 		         &8588 (48.6\%)		      	   & 1873	(10.6\%)	       \\
Total likes    & 53135 (2.9\%)              & 970478 (53.2\%)   		 & 133228 (7.3\%)       		 & 529039 (29.0\%)      	   &  139090 (7.6\%)   	   \\
Total comments & 796 (1.6\%)     			  & 28393 (56.1\%)           & 2488 (4.9\%)			         & 15231 (30.1\%)	           & 3670 (7.3\%)    	\\
Average likes     & 158.1      				  & 173.9    				 & 102.6          				 & 61.6       				   & 74.3          		   \\
Average comments   & 2.37     			  & 5.1			     & 1.92					     & 1.77 	 		       & 1.96           \\ 
\hline  

\multicolumn{1}{c}{\multirow{2}{*}{\begin{tabular}[c]{@{}c@{}} \textsf{Small Couture} \\ (N=6036, 70\% classified)  \end{tabular}}} \\ 
&Selfie & Body snap & Marketing & Product-Only & Non-fashion  \\ \hline
Total posts       & 35 (0.8\%)                 & 1092 (25.9\%)            & 365 (8.6\%)                     & 2543 (60.4\%)               & 183 (4.4\%)          \\
Total likes      & 39421 (7.7\%)              & 195256 (38.3\%)           & 29060 (5.7\%)                 & 234048 (45.9\%)              & 12262 (2.4\%)       \\
Total comments   & 220 (2.3\%)                 & 4273 (45.0\%)             & 586 (6.1\%)                   & 4192 (44.1\%)                & 246 (2.5\%)          \\
Average likes       & 1126.31              	   & 178.8               	 & 79.6                      	 & 92                     	   & 67                    \\
Average comments    & 6.29                  	 & 3.91                 	 & 1.61                      	& 1.65                   	   & 1.34                 \\
\hline
\multicolumn{1}{c}{\multirow{2}{*}{\begin{tabular}[c]{@{}c@{}} \textsf{High Street} \\ (N=5776, 71\% classified)  \end{tabular}}} \\ 
&Selfie & Body snap & Marketing & Product-Only & Non-fashion  \\ \hline 
Total posts       & 102 (2.4\%)                & 1725 (42\%)            & 376 (9.2\%)                   & 1376 (33.5\%)               & 532 (12.9\%)            \\
Total likes    & 3642 (0.9\%)                & 232937 (56.5\%)            & 4268 (10.3\%)                   & 67032 (16.2\%)                & 66442 (16.1\%)   \\
Total comments   & 150 (1.1\%)                & 8719 (66.3\%)            & 917 (7.0\%)                   & 2261 (17.2\%)                & 1122 (8.5\%)   \\
  Average likes       & 35.7                     & 135                 	 & 113.51                 		 & 48.72                       & 124.9        \\
Average comments    & 1.47                    	& 5.05           	         & 2.44                   	     & 1.64                  	   & 2.11           \\ 
\hline 
\end{tabular}
\vspace{-2mm}
\end{table*}

\subsection{Labeling Facial Emotions}

One of the prominent types in Table~\ref{imageclassification} is images containing human face(s), which take up 28\% of fashion posts. Faces are an important visual signal for evoking user engagement; a psychophysiological research found facial expression of happiness to be a major indicator of advertisement effectiveness~\cite{lewinski2014predicting}. In addition, photos with faces were seen to lead a higher degree of reactions on social media~\cite{bakhshi2014faces}.
In order to identify the importance of this factor, we consider 6,936 images that were classified as having a human face in our data and analyze them through the Microsoft's Emotion API\footnote{www.microsoft.com/cognitive-services/en-us/emotion-api}. The API gives out scores for the following kinds of emotions exhibited on a face: anger, contempt, disgust, fear, happiness, neutral, sadness, and surprise. The emotion API has been shown to have higher accuracy for classifying neutral and positive emotions than negative ones such as disgust~\cite{zhao2016emotion}.


We recognize the emotions of people by utilizing deep-learning based Microsoft Emotion API. According to their explanation, they identify emotions which are communicated cross-culturally and universally via the same basic facial expressions. This API detects up to 64 people for each image and the recognized faces are ordered by face size in descending order. The server responds with the results in a format which is human readable and easy to parse~\cite{schmidt2016cloud}.
The resulting face emotions are included in the labeled dataset. For each face, we list the results from eight emotion categories, where the highest score should be considered as the dominant emotion of that face.




\section{Research Findings}  

The labeled fashion data enable us to pursue a number of directions to better understand people's perception of fashion posts on social media. Here we list two such directions.

\subsection{Attention versus volume}

Given the five meaningful fashion categories, we examine how frequently each image type appears and how people reacted to such image type. Table~\ref{features_engagement} displays the relationship between the photo, likes, and comments counts. While the posting count was the highest for the product-only images that accounted for 34.6\% of all images followed by 22.5\% of images on body snaps, the average likes and comments counts were two to three times higher for the body snap images than the product-only images. The total proportion of likes and comments were disproportionately skewed toward the body snap and selfie images, compared to their post count proportions.  
 
The same analyses were conducted across brand groups. We show results for two of them due to space limitation. The number of product-only images tagging mega couture, small couture, and designer brands accounted for the largest fraction of all posts, whereas the average likes count of the body snap images was higher than those of the product-only images. A prominent pattern was further found in the average comments count. Selfies, while accounting for the smallest post count, gained high average likes count. However, in the case of high street brands, body snaps were the highest in volume even surpassing that of the product-only images. Body snaps were also the most liked, gaining more than twice the average likes count of product-only images. 

We provide possible explanation. First, high street brands are likely worn more comfortably by the public compared to expensive luxury goods. Hence, the volume of body snap images contributed by individuals may be the largest only for high street brands. When it comes to audience reaction, however, people seem to favor body snaps over product-only shots because such images of fashion products (appearing on everyday looks) are considered more informative. This trend appears across all brand groups, whether they are luxury or high street. Body snap fashion images are favored over the marketing and product-only images.
Second, selfies appear in fashion conversation although their focus is primarily on faces and not on fashion items. 
While one might hence assume selfies to be a less informative type of fashion conversation, it is noticeable to observe that selfies receive above the average likes and comments when they appear with luxury brand hashtags. Selfies that appear with high street brands in contrast were not equally popular. Why selfies with luxury brand tags are perceived as a positive signal (but not with hashtags of high street brands) is an interesting question to pursue in the future and this will help us understand how and why consumers use brands as symbols to create a favorable social identity~\cite{arnould2005consumer}.


\begin{table}[t!]
\centering
\caption{ANOVA test result on the likes count}
\label{ANOVA}
\begin{tabular}{lrrl} 
  \hline
 &  F value & Pr($>$F)\\ 
  \hline
Selfie &  1.459 &  0.2272\\ 
  Body &  51.512 & 7.32e-13&*** \\ 
  Marketing &   1.040 &  0.3079 \\ 
  Product-only &  5.362 &  0.0206&*   \\ 
  Non-fashion &  0.956 &  0.3281 \\  
  Face &  3.251 &  0.0714&. \\ 
  Logo & 15.811 & 7.02e-05&*** \\  \hline
  \multicolumn{4}{r}{.:p$<$0.1, *:p$<$0.05, ***:p$<$0.001}
 
\end{tabular}
\vspace*{-5mm}
\end{table}

\begin{figure}[b!]
\centering        \includegraphics[width=0.35\textwidth]{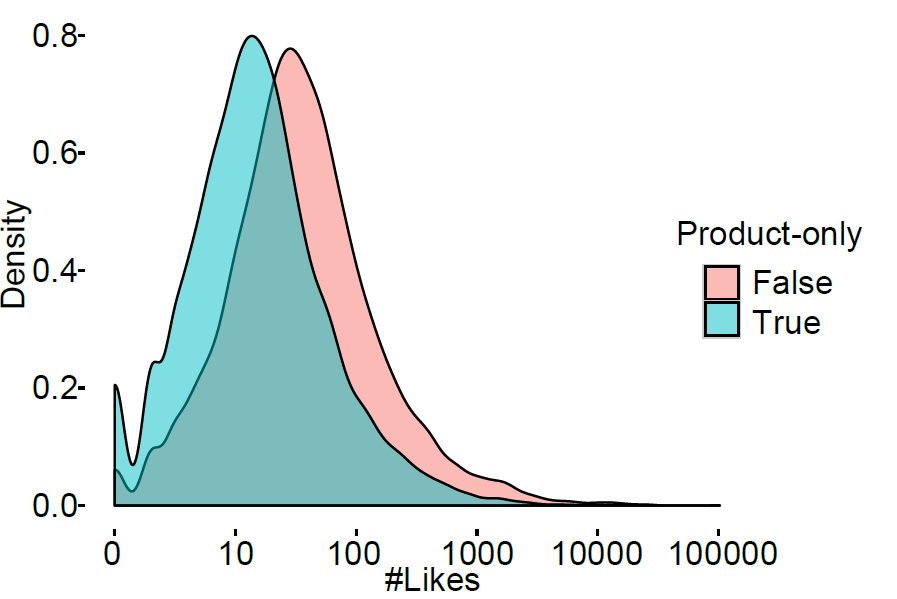}
        \vspace*{-3mm}
        \caption{Likes on images classified as product-only or not}
        \label{anova}
\end{figure}

We also examined the relationship between the visual content and the likes count via regression. For images involved in at least one visual category, we conducted one-way analysis of variance (ANOVA). Table~\ref{ANOVA} displays the result, indicating that four visual variables (body snap, product-only, face, and logo) have effects on the likes count. Body snap and face show a positive correlation, indicating that fashion images containing body parts or faces were more likely to attract likes from the audience. Note that the face feature is more general than selfies in that we do not restrict the size of a face in the former case. Product-only and logo had a negative correlation in that these features decreased the chance of acquiring extra likes. Figure~\ref{anova} demonstrates one such example on the product-only feature, where images classified as product-only show to be gaining less engagement from audience than otherwise. Existing literature ~\cite{manikonda2015trending} concludes that indirect product marketing (IM) which is photographed a fashion model holding the bag is more effective to the crowd than direct product marketing (DM). Our data also shows obvious product-only images appeal less favorably to Instagram audience than indirect marketing. 



\subsection{Fashionable faces}

As discussed earlier, face photos --- happy faces in particular --- are known to be effective in evoking user engagement. Our data match findings from previous work in that fashion images containing face(s) were notable; they were 28.0\% in volume, yet acquired 42.3\% and 35.7\% of all likes and comments, respectively. If faces are a critical visual feature in highly responded fashion images, what kinds of facial emotions are deemed fashion-likable? 




We adopt the robust regression model in order to test the influence of emotion scores on the likes count. The test utilized a log normalized count of likes  $log(\textrm{likes+1})$ as the dependent variable to account for heavy-tailed shape of the likes distribution. Table~\ref{lm} describes the fitted model, where two emotions from the Microsoft's Face API, fear and disgust, were omitted from results (as the tool states that these quantities are not fully tested in the current implementation).

\begin{table}[t!]
\centering
\caption{Robust regression results on the like count
}
\label{lm}
\begin{tabular}{rrrr}
  \hline
 & Estimate & Std. Error & t-value  \\ 
  \hline
  Anger & -0.0030 & 0.2616 & -0.0113  \\ 
  Contempt & 0.0972 & 0.3514 & 0.2766  \\ 
  Happiness & 0.3548  & 0.0210  & \textbf{16.8902}  \\ 
  Neutral & 0.3382 & 0.0148 & \textbf{22.9176}  \\ 
    Sadness & 0.5482 & 0.1852  & 2.9602  \\ 
    Surprise & 0.2713 & 0.1567 & 1.7311  \\ 
   \hline
\end{tabular}
\vspace*{-3mm}
\end{table}

In robust regression, a strong evidence is shown by large t-values instead of p-values. The table shows that two emotions, happiness and neutral, have positive influence on the likes count. Happy is expected from theory, yet neutral is an unexpected outcome. Our manual inspection revealed that images with neutral yet popular fashion posts exhibited  ``chic model-like faces'' that mimic fashion photography. The fact that this neutral non-smiling expression is prevalent not only by fashion models but also by individuals on Instagram in the context of fashion is worth noting and deems further investigation. 

\begin{figure}[t!]
        \centering        \includegraphics[width=0.4\textwidth]{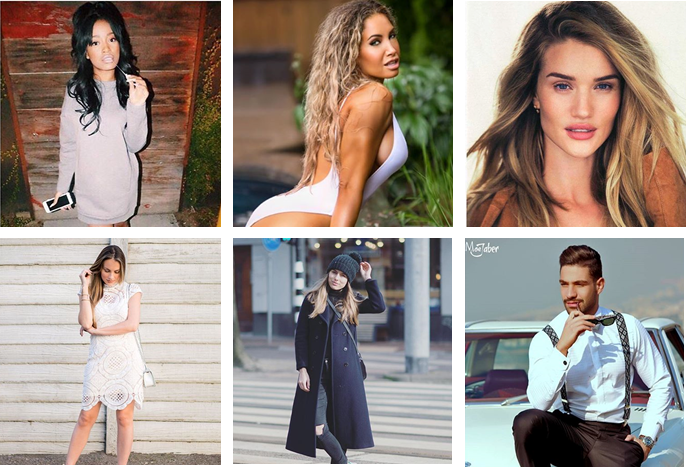}
        \vspace*{-1mm}
        \caption{Examples of popular fashion images that could not be classified into any of the eight classical face emotions }
        \label{emotion_failed}
        
\end{figure}

We found certain popular fashion images that received thousands of likes from audience to fail in showing any dominant emotion. Examples are shown in Figure~\ref{emotion_failed}, where none of the emotion scores were above 0.75. These six examples nonetheless had been responded well and had gained on average 4,915 likes. Despite the general tenancy that happy faces are predictive of likes, these examples demonstrate that new kinds of facial expressions need to be discovered to better understand characteristic emotions in fashion photography.

In fact, unidentified emotions were common in the data. Among fashion images that received more than 1000 likes, a non-negligible rate of 9.8\% of them could not be associated with any strong emotion based on existing tools. 
This finding indicates that there may be other kinds of emotions that are attractive in fashionable images.  These extraordinary cases shared in our data could be used as a valuable and unique machine learning training set for identifying facial expressions that attract fashion-avid people. 




 \section{Discussion and Conclusion}

\subsubsection{Data sharing contribution}
 
Visual-centric social media have become a key marketing playground for fashion industry by serving as a major communicative channel between brands and customers. Although this transformation is delivering a tremendous amount of impact, little has been systematically studied due to the \textit{lack} of any relevant dataset. Gathering and sharing  fashion data would enable the kinds of analysis that we see in many other domains. This paper presents a novel dataset of fashion posts on Instagram and this is the first sizable dataset specifically focused on and tailored to the studies of fashion.

Our fashion dataset comprises 
i) description of images (e.g., hashtags, text description), ii) associated metadata (e.g., date, likes count), and iii) annotated and predicted visual content variables. Therefore, the data can be used to support various studies in social media analysis with a unique dimension of visual content and its role in online viral marketing, brand promotions, and user engagement. 
 

Through an iterative procedure of manual tagging and reviewing, we discovered major visual signals that can effectively characterize fashion post styles and frequent components of fashion photographs. To efficiently scale up the size of our dataset, we trained a CNN-based image classifier on the annotated data and applied the model to the entire image set. A quantitative evaluation was performed to validate the accuracy of the model. Furthermore, we adopted a public software to classify the emotional status of faces in images, which provides a subtle, sentimental cue at a deeper level. 

\subsubsection{Implications of findings}

Our key findings can be summarized as follows. First, the most common image type was `product-only' although the `body snap' or `selfie' images were more effective in eliciting user engagement such as likes or comments. Such ``natural'' fashion exposures were more frequently found among high street brands. Second, we qualitatively reviewed several failure cases in tagging to identify challenges and weakness of the current analysis and discovered a few common types. These types include advertisements, clickbaits and zoomed-in images of textile. While some images were not related to fashion at all, there exist images which do not fall into the considered categories yet still provide useful information about fashion products. An issue that can be addressed in the future is therefore to incorporate a larger number of visual variables on fashion images.

\begin{figure}[t!]
	\centering        \includegraphics[width=0.45\textwidth]{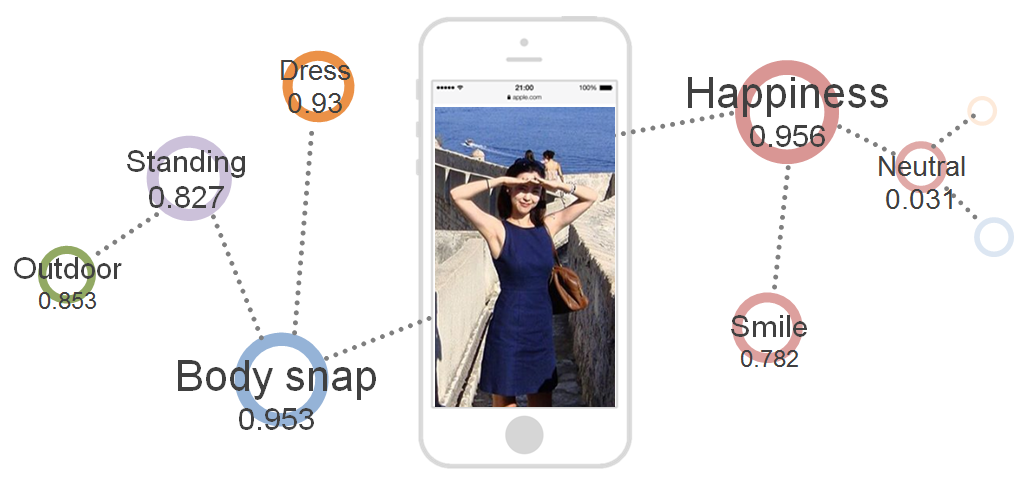}
    \vspace*{-2mm}
	\caption{An example of possible application based on our dataset. Our labeled data can be utilized to recommend users to join peer groups who have similar fashion taste.}
	\label{Application}
    \vspace*{-3mm}
\end{figure} 


\subsubsection{Potential applications}

There are many outcomes that one may produce from analyzing the fashion data. Here we envision one such application in 
Figure~\ref{Application}, which analyzes fashion images for fashion-avid individuals, brand marketers, and fashion experts. Upon receiving or reading in a fashion image update, the system can analyze fashion-relevant visual features such as image category (e.g., body snap vs product-only), face emotion (e.g., happy vs chic), visual contexts (e.g., outdoor vs indoor), as well as other cues (e.g., posture, color). Then the system can match fashion posts composed of similar characteristics yet received more likes and comments. Such a system can help individuals, marketers, and brand experts recognize appealing styles, competing brands, and upcoming trends. Existing tools, however, cannot immediately provide such service, because they do not know 1) what signals are meaningful and dominant, 2) what visual features are more appealing, 3) which images should be considered similar in the context of fashion conversation. The data shared in this paper allows for such investigation.



Beyond envisioning an app service, fashion datasets  enable us to study how fashion tastes are shared on social media, each season and across the globe. The ability to track new trends at an unprecedented scale will be critical to capturing how fast fashion trends turn around. 
%
We are also interested in knowing how fashion purchases are determined, to what extent the growing Internet population directly or indirectly learn about fashion trends through online platforms. Finally, we would like to learn more about how the collective online experience (whether seeing information posted by brands or individuals) contributes to building brand images, which affect future purchase intent. 

\subsubsection{Limitations and future directions}
This study bares several limitations. First, due to small data, our research cannot bring out aspects of diverse cultural differences on fashion. We plan to gather data across multiple cultural domains for comparison. Second, several important elements of fashion such as zoom-in images of textile or the color of fashion items were not handled in our visual cues. Future studies can systematically define these fashion feature types and build algorithms to classify such features from fashion images. Third, we did not distinguish patterns from the official brands themselves and individuals. One may consider other meaningful groups such as micro-celebrity, fashion retailers, and grassroots. Analyzing conversations by  user groups will be helpful in classifying fashion information more concretely. By taking the initiative to prepare and share sizable fashion data, we hope our effort is joined by many other fashion researchers and brand experts who are interested in discovering fashion tastes.

\section{
Acknowledgement}

{
Cha and Ha were supported by the Ministry of Trade, Industry \& Energy (MOTIE, Korea) under Industrial Technology Innovation Program (No.10073144), `Developing machine intelligence based conversation system that detects situations and responds to human emotions'. For correspondence on findings and dataset, please contact the third author at meeyoungcha@gmail.com.}

\bibliographystyle{aaai}
\bibliography{reference}

\end{document}